\PassOptionsToPackage{utf8}{inputenc}
\documentclass{bioinfo}
\copyrightyear{2015} \pubyear{2015}

\access{Advance Access Publication Date: Day Month Year}
\appnotes{Manuscript Category}
\usepackage{lmodern}
\usepackage{subfigure}
\usepackage{url}
\usepackage{color}
\usepackage{multirow,booktabs}
\usepackage{float}
\begin{document}
\firstpage{1}

\subtitle{Subject Section}

\title[short Title]{HampDTI: a heterogeneous graph automatic meta-path learning method for drug-target interaction prediction}
\author[Sample \textit{et~al}.]{Hongzhun Wang\,$^{\text{\sfb 1,}}$, Feng Huang\,$^{\text{\sfb 1,}}$, Wen Zhang\,$^{\text{\sfb 1}*}$}
\address{$^{\text{\sf 1}}$College of Informatics, Huazhong Agricultural University, Wuhan 430070, China}

\corresp{$^\ast$To whom correspondence should be addressed.}

\history{Received on XXXXX; revised on XXXXX; accepted on XXXXX}

\editor{Associate Editor: XXXXXXX}

\abstract{\textbf{Motivation:} Identifying drug-target interactions (DTIs) is a key step in drug repositioning. In recent years, the accumulation of a large number of genomics and pharmacology data has formed mass drug and target related heterogeneous networks(HNs), which provides new opportunities of developing HN-based computational models to accurately predict DTIs. The HN implies lots of useful information about DTIs but also contains irrelevant data, and how to make the best of heterogeneous networks remains a challenge.\\
\textbf{Results:} In this paper, we propose a heterogeneous graph automatic meta-path learning based DTI prediction method (HampDTI). HampDTI automatically learns the important meta-paths between drugs and targets from the HN, and generates meta-path graphs. For each meta-path graph, the features learned from drug molecule graphs and target protein sequences serve as the node attributes, and then a node-type specific graph convolutional network (NSGCN) which efficiently considers node type information (drugs or targets) is designed to learn embeddings of drugs and targets. Finally, the embeddings from multiple meta-path graphs are combined to predict novel DTIs. The experiments on benchmark datasets show that our proposed HampDTI achieves superior performance compared with state-of-the-art DTI prediction methods. More importantly, HampDTI identifies the important meta-paths for DTI prediction, which could explain how drugs interact with targets in HNs.\\
\textbf{Contact:} \href{name@bio.com}{name@bio.com}\\
\textbf{Supplementary information:} Supplementary data are available at \textit{Bioinformatics}
online.}

\maketitle

\section{Introduction}
The study \citep{Paul2010HowTI} indicates that each new molecular entity discovery costs about \$1.8 billion and the approval of new drug applications takes on average 9 to 12 years. Drug repurposing offers a low-cost alternative approach, where researchers labour for finding new therapeutic targets for approved drugs. In the era data explosion, in silico predicting drug-target interactions (DTIs) promises to significantly accelerate repositioning of drugs \citep{keiser2009predicting, wu2017sdtnbi}. Therefore, it is necessary to develop efficient computational methods which could accurately predict DTIs with high reliability for further experiments. Traditional computational methods can be mainly divided into two categories, including the docking-based methods\citep{li2006tarfisdock} and the ligand-based methods\citep{keiser2007relating}. However, the performance of ligand-based methods is poor when the target proteins have few known ligands, while the docking-based methods are extremely dependent on the three-dimensional structure of the proteins, and the performance is limited when the three-dimensional structures of the target proteins are unknown.

To overcome above mentioned limitations, a number of machine learning-based approaches have been developed to achieve accurate DTI prediction. The similarity-based approaches take up the majority in the machine learning-based DTI prediction, which assumes that similar drugs may share similar targets and vice versa. They are further categorized into kernel-based approaches\citep{van2011gaussian}, matrix factorization-based approaches\citep{zheng2013collaborative,ezzat2016drug,liu2016neighborhood} and network-based approaches\citep{mei2013drug,cheng2012prediction}. In particular, network-based approaches have showed great advantages and attracted more and more attention in recent years.

With the development of life science as well as pharmaceutical technology, biomedical data resources have been rapidly accumulated to form drug and target related heterogeneous networks (HNs) with various bioentities and associations, which provides more diverse information for DTI prediction. The manifold learning\citep{chen2012drug,cai2016multi}, matrix factorization\citep{wang2014drug} and random walk\citep{chen2012drug,wang2021bioerp} are employed to HN-based DTI prediction. These methods effectively improve the accuracy of DTI prediction by integrating diverse information from heterogeneous networks. Recently, graph neural networks (GNNs) which apply deep learning on graphs have been developed and have demonstrated good performances on various network-related tasks\citep{li2020neural,yu2021predicting}. Several GNN-based models have been proposed for DIT prediction from the HN. For example, \cite{NeoDTI} introduces relational graph convolutional networks (R-GCNs) to the HN for DTI prediction; \cite{GCN-DTI} first constructs a homogeneous DPP (drug protein pair) network from the drug and target related HN, then uses graph convolutional network (GCN) to the DPP network.

Although the above HN-based methods produce promising results on DTI prediction, they neglect the semantic meaning of different meta-paths, which is crucial for analyzing HNs. \cite{fu2016predicting} reports that meta-path-based topological features could significantly improved the HN-based DTI prediction performance. \cite{wang2021bioerp} uses random walk guided by meta paths, which has achieved  performance boost in several biological HN relationship prediction tasks, especially in DTI prediction. The drug and target related HN contain various bio-entities, where drugs and targets are connected via different meta-paths, but not all meta-paths contribute to DTI prediction. \cite{fu2016predicting}, \cite{wang2021bioerp} and most of other existing studies manually select meta-paths that are forecast to benefit DTI prediction. Unfortunately, experience-based selection of meta-paths relies on domain knowledge and is not always adaptable and transferable across different HNs. Therefore, there is a requirement to explore an automatic meta-path learning method for heterogeneous network-based DTI prediction.

In this paper, we propose a heterogeneous graph automatic meta-path learning based DTI prediction method HampDTI, which not only utilizes heterogeneous network-derived semantic information but also considers molecular structure and protein sequence information. First of all, HampDTI extracts features of drugs and proteins from drug molecule graphs and protein sequences. Meanwhile, HampDTI automatically learns meta-paths between drugs and targets from the heterogeneous graph, and generates meta-path graphs. Then, the learned drug and target features are taken as the node attributes in the meta-path graphs, and a node-type specific graph convolutional network (NSGNN) is designed to learn node embeddings from the attributed meta-path graphs. Finally, HampDTI combines the embeddings from all meta-path graphs to predict novel DTIs. The main contributions are summarized as follows:

\begin{itemize}
\item For drug and target related heterogeneous networks, we design an automatic meta-path learning method to learn the important meta-paths between drugs and targets. 

\item For each meta-path graph, we devise a node-type specific graph convolutional network (NSGNN) which considers the information of different node types(drugs and targets).

\item The proposed DTI prediction method HampDTI integrates drug  molecule  structure  and  target  protein  sequence information as well as direct and indirect drug-target relations within  the  learned  meta-paths,  which improves the accuracy of DTI prediction.
\end{itemize}

\begin{methods}
\section{Methods}
\subsection{Problem Definition}

\hspace{1em} \textit{Definition 2.1} \textbf{Heterogeneous Graph}. A drug and target related heterogeneous graph is defined as a graph $\mathcal{G}=(\mathcal{V}, \mathcal{E})$ with a node type mapping function $\phi: \mathcal{V} \rightarrow \mathcal{T}^v$ and an edge type mapping function $\psi: \mathcal{E} \rightarrow \mathcal{T}^e$. $\mathcal{T}^v$ and $\mathcal{T}^e$ denote the predefined sets of node types and edge types, where $|\mathcal{T}^v| + |\mathcal{T}^e|>2$ and $\mathcal{T}^v$ includes drug, target and other bio-entities. \vspace{10pt}

\textit{Definition 2.2} \textbf{Meta-path}. In the heterogeneous graph $\mathcal{G}$, a meta-path $P$ is a relation sequence between two node types. i.e., $v_{1} \stackrel{e_{1}}{\rightarrow} v_{2} \stackrel{e_{2}}{\rightarrow} v_{3} \cdots \cdots \stackrel{e_{p}}{\rightarrow} v_{p+1}$ denotes a meta-path of length $p$, where $e_i \in \mathcal{T}^e$ and $v_{i} \in \mathcal{T}^v$. \vspace{10pt}

\textit{Definition 2.3} \textbf{Meta-path Graph}. Given a meta-path $P$ of the heterogeneous graph $\mathcal{G}$, all node pairs connected via meta-path $P$ construct the meta-path graph $\mathcal{G}^P$. \vspace{10pt}

We formulate the DTI prediction as a classification task to determine whether one drug-target interaction is present or not. Given drug SMILES $\mathcal{U}=\left\{u_{1}, \cdots, u_{m}\right\}$ for $m$ drugs, protein sequences $\mathcal{S}=\left\{s_{1}, \cdots, s_{n}\right\}$ for $n$ proteins and the heterogeneous graph $\mathcal{G}=(\mathcal{V}, \mathcal{E})$, the DTI prediction task can be seen as learning a function $\mathcal{F}(\mathcal{U},\mathcal{S},\mathcal{G}) \to [0,1]$ to compute the probability of interactions between drugs and targets in the heterogeneous graph $\mathcal{G}$.


\subsection{The HampDTI Method}
Figure \ref{fig01} illustrates the workflow of the HampDTI. First, the \textbf{feature learning module} learns drug structure features and protein sequence features from SMILES and protein sequences. Meanwhile, the \textbf{automatic meta-path learning module} automatically learns meta-paths between drugs and proteins from heterogeneous graph and constructing meta-path graphs with drug structure features and protein sequence features as the node attributes. Then, the \textbf{node-type specific graph convolutional network module} learns node embeddings from each meta-path graph. Finally, the \textbf{embeddings incorporation and prediction module} incorporates multiple embeddings from different meta-path graphs to achieve accurate DTI prediction.

\begin{figure*}[!htbp]
\centering
\includegraphics[width=2\columnwidth]{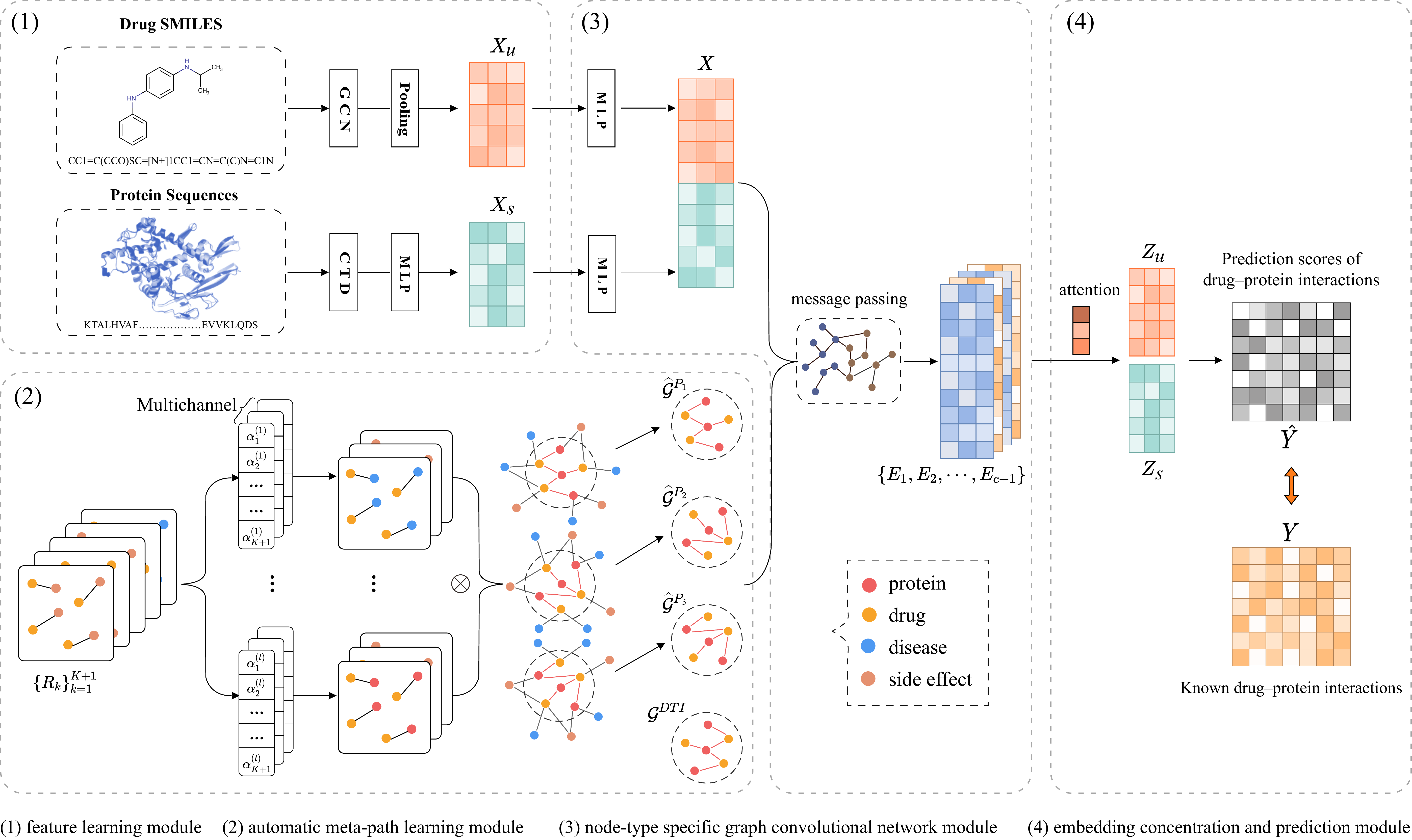}
\caption{The schematic workflow of the HampDTI. (1) feature learning module: learning drug structure features $X_u$ and protein sequence features $X_s$ from SMILES and protein sequences. (2) automatic meta-path learning module: learning meta-paths between drugs and proteins from heterogeneous graph and constructing meta-path graphs. (3) node-type specific graph convolutional network module: learning node embeddings from multiple meta-path graphs. (4) embedding concentration
and prediction module: incorporating multiple embeddings to predict novel DTIs.}
\label{fig01}
\end{figure*}


\begin{table*}[!htbp]
\centering
\caption{Atomic features}
\setlength{\tabcolsep}{10mm}{
\begin{tabular}{lp{30em}l}
\toprule  
Number& Feature& Dimension\\
\midrule  
1& One-hot encoding of the atom element& 44\\
2& One-hot encoding of the degree of the atom in the molecule,which is the number of directly-bonded neighbors(atoms)& 11\\
3& One-hot encoding of the total number of H bound to the atom& 11\\
4& One-hot encoding of the number of implicit H bound to the atom& 11\\
5& Whether the atom is aromatic& 1\\
 & All& 78\\
\botrule 
\end{tabular}}
\label{tab01}
\end{table*}

\subsubsection{Feature learning module}
The feature learning module learns drug structure features and protein sequence features from SMILES and protein sequences. The drug SMILES $\left\{u_{1}, u_{2}, \cdots, u_{m}\right\}$ are converted into drug molecule graphs using Rdkit \citep{Landrum2016RDKit2016_09_4}, with each atom as a node and chemical bonds as edges. For drug molecule graphs, GCN \citep{kipf2016semi} can be applied to encode drug molecule structure into a graph-level representation as drug structure feature. Formally, for the $i^{th}$ drug, the updated node representation in the $l^{th}$ layer is defined as:

\begin{equation}
H^{(l)}_i=\sigma\left(\tilde{D}^{-\frac{1}{2}}_i \tilde{A}_i \tilde{D}^{-\frac{1}{2}}_i H^{(l-1)}_i W^{(l)}\right)
\label{eq01}
\end{equation}

\noindent where $\tilde{A}_i = A_i + I$ is the adjacency matrix of the $i^{th}$ drug molecule graph with added self-connections, $\tilde{D}_i$ is the degree matrix of $\tilde{A}_i$, $W^{(l)}$ is a weight matrix and $\sigma$ is a activation function. Specially, $H^{(0)}_i \in \mathbb{R}^{q_i \times 78}$ is the $i^{th}$ drug atomic feature matrix (the atomic features are showed in Table \ref{tab01}), where $q_i$ denotes the number of atoms contained in the $i^{th}$ drug.

Fllowed by $L$ layers GCN, a global max pooling layer is applied to obtain the $i^{th}$ drug structure feature $X_{u(i)} \in \mathbb{R}^{d} = pool(H^{(L)}_i)$, where $H^{(L)}_i \in \mathbb{R}^{q_i \times d}$ and $d$ is the feature dimension. For $m$ drugs, $X_u \in \mathbb{R}^{m \times d}$ denotes all drug structure features learned from the molecule graphs of all the $m$ drugs.

For protein sequences $\left\{s_{1}, s_{2}, \cdots, s_{n}\right\}$, the Composition, Transition and Distribution (CTD) \citep{dubchak1995prediction,dubchak1999recognition} is applied to convert each sequence into an 147-dimensional feature vector. Then multi-layer perceptron (MLP) is applied to obtain the compact protein sequence features $X_s \in \mathbb{R}^{n \times d}$ and unifies the dimension of drug structure features and protein sequence features.


\subsubsection{Automatic meta-path learning module}
To find most important meta-paths for DTI prediction, we designed automatic meta-path learning module based on GTN \citep{NEURIPS2019_9d63484a}. The automatic meta-path learning module learns meta-paths between drugs and proteins in the heterogeneous graph $\mathcal{G}$.

The heterogeneous graph $\mathcal{G}=(\mathcal{V}, \mathcal{E})$ can be represented by a set of adjacency matrices $\mathcal{R} = \{R_k\}_{k=1}^K$ where $K=|\mathcal{T}^e|$ and $R_k \in \mathbb{R}^{|\mathcal{V}| \times |\mathcal{V}|}$ is an adjacency matrix where $R_k[i, j]$ is non-zero when there is a $k^{th}$ type edge from node $i$ to $j$. Given a mata-path $P: v_{1} \stackrel{e_{1}}{\rightarrow} v_{2} \stackrel{e_{2}}{\rightarrow} v_{3} \cdots v_{i} \stackrel{e_{i}}{\rightarrow} \cdots \stackrel{e_{p}}{\rightarrow} v_{p+1}$, we can build a meta-path graph $\mathcal{G}^P$, and the adjacency matrix $A_{\mathcal{G}^P}$ of the meta-path graph $\mathcal{G}^P$ is obtained by the multiplications of the corresponding adjacency matrices as:

\begin{equation}
A_{\mathcal{G}^P}=R_{e_1}R_{e_2} \cdots R_{e_i} \cdots R_{e_p}
\label{eq03}
\end{equation}

\noindent where $R_{e_i}$ is the adjacency matrix corresponds to edge type $e_i$. 

To generate one meta-path graph corresponding to a $p$-length meta-path, we can select $p$ adjacency matrices $\{ R_{e_1},R_{e_2}, \cdots, R_{e_p} \}$ from $\mathcal{R}$ with replacement and multiply them as Eq. (\ref{eq03}) to obtain the adjacency matrix of the meta-path graph. To achieve automatic meta-path learning, we softly select $\hat{R}_{e_i}$ from $\mathcal{R}$ by computing the convex combination of adjacency matrices as:

\begin{align}
\hat{R}_{e_i} = \sum_{k=1}^{K} \alpha_k^{(i)} R_k , \ i = 1,2, \cdots , p \label{eq04}
\end{align}

\noindent where $\alpha_k^{(i)}$ is a weight for relation $R_k$ in $i^{th}$ soft selection, and it is defined as:

\begin{equation}
\alpha_k^{(i)} = \frac{\mathrm{exp}(e_k^{(i)})}{\sum_{k=1}^{K} \mathrm{exp}(e_k^{(i)})} 
\end{equation}

\noindent where $e_k^{(i)}$ is a learnable parameter. Then we can obtain an adjacency matrix $\hat{A}_{\mathcal{G}^P} = \hat{R}_{e_1}  \hat{R}_{e_2}  \cdots  \hat{R}_{e_p}$ by Eq. (\ref{eq03}). The $\hat{A}_{\mathcal{G}^P}$ is learnable which means we can automatically learn meta-paths. 

The $\hat{A}_{\mathcal{G}^P} \in \mathbb{R}^{|\mathcal{V}| \times |\mathcal{V}|})$ contains all node of the heterogeneous graph $\mathcal{G}$. In order to accurately predict the novel DTI, we only focus on the meta-paths between drug and target. Therefore, we extracts subgraphs $\mathcal{\hat{G}}^P (\hat{A}_{\mathcal{\hat{G}}^P} \in \mathbb{R}^{m \times n})$ only containing drug entities and target entities from the meta-path graph $\mathcal{G}^P (\hat{A}_{\mathcal{G}^P} \in \mathbb{R}^{|\mathcal{V}| \times |\mathcal{V}|})$.

To learn multiple meta-paths simultaneously, we use $c$ channels in parallel to obtain multiple meta-path graphs $\{\hat{\mathcal{G}}^{P_1}, \hat{\mathcal{G}}^{P_2}, \cdots ,\hat{\mathcal{G}}^{P_c}\}$. Specially, we add the identity matrix $I$ to $\mathcal{R}$ for learning variable-length meta-paths. This trick allows us to learn any length of meta-paths up to $p$ when we execute $p$ soft selections. Therefore $p$ in Eq. (\ref{eq04}) denotes the maximum length of the learned meta-paths.


\subsubsection{Node-type specific graph neural network module}
To learn satisfactory node embeddings from meta-path graphs, we designed node-type specific graph convolutional network(NSGCN). Most existing GNNs follow a paradigm of entangling feature transformation and message propagation, which leads to a large number of parameters and is prone to over-smoothing issue \citep{liu2020towards}. Therefore NSGCN separates feature transformation and message propagation operation to alleviate these issue. In addition, we utilizes two different MLPs for drug nodes and target nodes to effectively consider node type information in the feature transformation stage. The feature transformation is defined as:

\begin{align}
X = [MLP_u\left(X_{u}\right);MLP_s\left(X_{s}\right)] \label{eq10}
\end{align}

\noindent where $X_u$ is drug features and $X_s$ is protein features mentioned in section 2.2.1. The $X_u$ and $X_s$ are used as the initial node features of the meta-path graphs.

Then the message propagation in $t^{th}$ layer is defined as:

\begin{align}
E_i=&{(D^{-\frac{1}{2}}(\hat{A}_{\mathcal{\hat{G}}^{P_i}}+I) D^{-\frac{1}{2}})}^t X \label{eq11}\\
&i=1,2, \cdots ,c+1 \nonumber   
\end{align}

\noindent where $E_i$ is node embeddings learned from the $i^{th}$ meta-path graph $\mathcal{\hat{G}}^{P_i}$, $\hat{A}_{\mathcal{\hat{G}}^{P_i}}$ is  the adjacency matrix of $\mathcal{\hat{G}}^{P_i}$ and $c$ is the number of learned meta-path graphs. In addition to the meta-path graphs, we extra use the drug-target interaction graph $\mathcal{G}^{DTI}$  to consider the direct interaction information between the drugs and the targets. Given the meta-path graphs $\{\hat{\mathcal{G}}^{P_1}, \hat{\mathcal{G}}^{P_2}, \cdots ,\hat{\mathcal{G}}^{P_c}\}$ and drug-target interaction graph $\mathcal{G}^{DTI}$, NSGCN learns node embeddings $\{E_1, E_2, \cdots , E_{c+1}\}$ from each graph independently.


\subsubsection{embedding concentration and prediction module}
To obtain more comprehensive node embeddings, we use attention mechanism to incorporate multiple embeddings $\{E_1, E_2, \cdots , E_{c+1}\}$ which indicate multiple semantics of the heterogeneous graph. The attention can automatically learn the importance of different node embeddings from different meta-path graphs. we fuse these embeddings to obtain the final embedding $Z$ as follows:

\begin{align}
Z &= \sum_{i=1}^{c+1} \frac{\mathrm{exp}(a_i)}{\sum_{i=1}^{c+1} \mathrm{exp}(a_i)} E_i
\end{align}

\noindent where $a_i$ is a learnable parameter.

The drug embeddings $Z_{u}$ and protein embeddings $Z_{s}$ are obtained from the final embedding $Z$. Then the reconstructed DTI matrix can be written as:

\begin{equation}
\hat{Y}=Z_{u}\ Z_{s}^T
\label{eq55}
\end{equation}

Finally, the HampDTI model can be optimized through the follow loss function:

\begin{equation}
\ell =  (1-\gamma) || Y \odot (Y-\hat{Y}) ||^2 + \gamma || (1-Y) \odot (Y-\hat{Y}) ||^2
\label{eq15}
\end{equation}

\noindent where $\odot$ is the element-wise product, $Y$ is the ground truth label, $|| \cdot ||^2$ is the squared Frobenius norm and $\gamma$ is a hyperparameters.

\end{methods}

\section{Result}
\subsection{Dataset}
In this work, we adopted the DTINet dataset curated by previous study\citep{DTInet}. As shown in Table \ref{tab02}, the DTINet dataset includes six drug/protein related networks: drug-protein interaction network, drug-drug interaction network, protein-protein interaction network, drug-disease association network, protein-disease association network and drug-side effect association network. Additionally, for all drug entities and protein entities in the drug-protein interaction network, we collected SMILES of drugs from Drugbank\citep{wishart2018drugbank} and sequences of proteins from Uniprot\citep{uniprot2019uniprot}.

\begin{table}[!htbp]
\centering
\caption{statistics of DTINet}
\label{tab02}
\setlength{\tabcolsep}{9pt}{
\begin{tabular}{lrlr}
\toprule  
Node Types& Num& Edge Types& Num\\
\cmidrule(lr){1-2}\cmidrule(lr){3-4}  
Drug& 708& Drug-Protein& 1,923\\
Protein& 1,512& Drug-Drug& 10,036\\
Disease& 5,603& Protein-Protein& 7,363\\
Side Effect& 4,192& Drug-Disease& 199,214\\
 & & Protein-Disease& 1,596,745\\
  & & Drug-Side Effect& 80,164\\
Total& 12,015& Total& 1,895,445\\
\botrule 
\end{tabular}}
\end{table}

\begin{figure*}[!htbp]
\centering
\includegraphics[width=1.8\columnwidth]{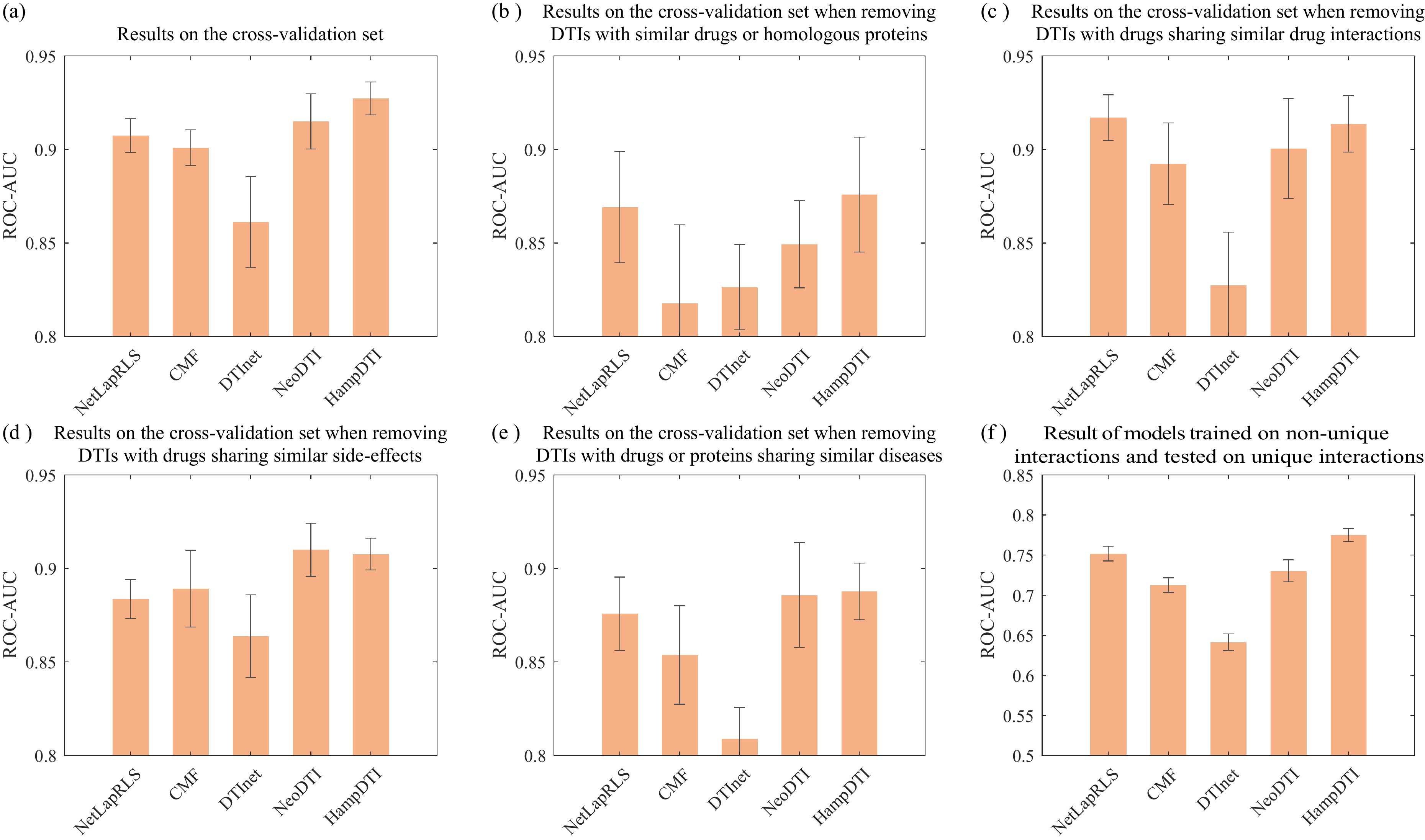}
\caption{The performance evaluation results of HampDTI on several challenging scenarios in terms of ROC-AUC scores. In each scenario, we performed a 10-fold cross-validation to compare with the baseline. (a)results on the cross-validation set. (b)results on the cross-validation set when removing DTIs with similar drugs or homologous proteins. (c)results on the cross-validation set when removing DTIs with drugs sharing similar drug interactions. (d)results on the cross-validation set when removing DTIs with drugs sharing similar side-effects. (e)results on the cross-validation set when removing DTIs with drugs or proteins sharing similar diseases. (f) results of models trained on non-unique interactions and tested on unique interactions.}
\label{figcompare}
\end{figure*}

\begin{table*}[!htbp]
\centering
\caption{Performances evaluated on cross-validation set}
\label{tab04}
\setlength{\tabcolsep}{4mm}{
\begin{tabular}{lccccccc}
\toprule  
 & ROC-AUC & PR-AUC & $F_1$ & Accuracy & Recall & Specificity & Precision\\
\midrule
NetLapRLS   &0.9074 &0.9180 &0.8388 &0.8321 &0.8730 &0.7911 &0.8094 \\
CMF         &0.9010 &0.9247 &0.8468 &0.8487 &0.8367 &\textbf{0.8606} &\textbf{0.8585} \\
DTInet      &0.8612 &0.8861 &0.7996 &0.8029 &0.7853 &0.8205 &0.8263 \\
NeoDTI      &0.9150 &0.9173 &0.8562 &0.8550 &0.8628 &0.8472 &0.8505 \\
HampDTI   &\textbf{0.9273} &\textbf{0.9263} &\textbf{0.8689} &\textbf{0.8648} &\textbf{0.8986} &0.8310 &0.8431 \\
\botrule 
\end{tabular}}
\end{table*}

\begin{table*}[!htbp]
\centering
\caption{Performances evaluated on independent test set}
\label{tabindepent}
\setlength{\tabcolsep}{4mm}{
\begin{tabular}{lccccccc}
\toprule  
 & ROC-AUC & PR-AUC & $F_1$ & Accuracy & Recall & Specificity & Precision\\
\midrule
NetLapRLS   &0.9003 &0.8993 &0.8506 &0.8515 &0.8469 &0.8562 &0.8554 \\
CMF         &0.8970 &\textbf{0.9210} &0.8461 &0.8487 &0.8313 &\textbf{0.8661} &\textbf{0.8639} \\
DTInet      &0.8505 &0.8567 &0.7928 &0.7823 &0.8333 &0.7313 &0.7564 \\
NeoDTI      &0.9057 &0.9002 &0.8518 &0.8458 &0.8839 &0.8078 &0.8235 \\
HampDTI      &\textbf{0.9209} &0.8972 &\textbf{0.8824} &\textbf{0.8776} &\textbf{0.9193} &0.8359 &0.8491 \\
\botrule
\end{tabular}}
\end{table*}

\subsection{Experiment settings}
In our experiments, we treated the drug-target pairs with known interactions as positive samples and the remaining drug-target pairs as negative samples. A randomly chosen subset of 10\% positive samples and a matching number of randomly sampled negative samples were held out as the independent test set. The remaining samples were used as the cross-validation set. The cross-validation set is used to implement 10-fold cross-validation experiments and optimize the hyper-parameters. The independent test set is adopted for the in-deep comparison between the proposed method and other baseline methods. We considered several evaluation metrics: ROC-AUC (area under the receiver operating characteristic curve), PR-AUC (area under the precision-recall curve), F1 score, Accuracy, Recall, Specificity and Precision.

The proposed method HampDTI has four important hyper-parameters: (\romannumeral1) the dimension of the node embedding $d$; (\romannumeral2) the coefficient $\gamma$ in Eq. (\ref{eq15}); (\romannumeral3) the maximum length of the learned meta-paths $p$; (\romannumeral4) the number of learned meta-paths $c$. To tuning hyper-parameters, we consider $d \in \{128, 256,512\}$, $\gamma \in \{0.3, 0.4, 0.5, 0.6 0.7\}$, $p \in \{2, 3, 4\}$, and $c \in \{3, 4, 5\}$. Finally, we set hyper-parameters $d=128, \gamma=0.4, p=3, c=4$ according to  10-fold cross-validation results.

Our model training process different from the conventional end-to-end training paradigm, we independently train the feature learning module using the same loss function in Eq. (\ref{eq15}), then train the remaining modules.

\subsection{Comparison with other state-of-the-art methods}
\subsubsection{Baselines}
We compared the performance of HampDTI with the following state-of-the-art network-based methods for DTI prediction, including NetLapRLS\citep{xia2010semi}, CMF\citep{zheng2013collaborative}, DTINet\citep{DTInet} and NeoDTI\citep{NeoDTI}.

\textbf{NetLapRLS} is a manifold regularization semi-supervised learning method for drug-protein interaction prediction by integrating information from drug-protein interaction network, drug-drug similarity network and protein-protein similarity network.

\textbf{CMF} is a multi-similarity cooperative matrix decomposition algorithm, which maps drugs and proteins into a common low-dimensional space. It integrates more diverse similarity network information compared to NetLapRLS.

\textbf{DTINet} constructs a drug/protein related heterogeneous network dataset: DTINet dataset. It uses RWR ( Random Walk with Restart ) and DCA ( Diffusion Component Analysis ) algorithm to obtain a low-dimensional vector representation of drugs and proteins.

\textbf{NeoDTI} employs a graph neural network to learn topology-preserving node embeddings by reconstructing the edge weight of the heterogeneous network. 

For all the compared methods, their hyper-parameters are determined as discussed in their corresponding papers. the embedding dimension of drugs and targets in DTINet, NeoDTI and HampDTI are all set to 128.

\subsubsection{Performances evaluated by 10-fold cross-validation}
Here, we implemented the 10-fold cross-validation to evaluate performances of our method HampDTI and other state-of-the-art methods on cross-validation set. As shown in Table \ref{tab04},  HampDTI outperforms the state-of-the-art methods, improving 1.7\% in terms of ROC-AUC and 0.2\% in terms of PR-AUC compared to the second best method.

The DTINet dataset contains homologous proteins and similar drugs, which can lead to overstated performance of the model by easy predictions. For further study, we conducted additional four tests on cross-validation set which follow the same strategy in \cite{DTInet}: (i)  removing DTIs with similar drugs (drug chemical structure similarities >0.6) or homologous proteins (protein sequence similarities > 0.4); (ii) removing DTIs with drugs sharing similar drug interactions (Jaccard similarities >0.6); (iii) removing DTIs with drugs sharing similar side-effects (Jaccard similarities >0.6); (iv) removing DTIs with drugs or proteins sharing similar diseases (Jaccard similarities >0.6), and the results are shown in 
Figure \ref{figcompare} (b-e) respectively. For comparison, we also demonstrate the results without removing similarity in Figure \ref{figcompare} (a), and HampDTI produces decreased performances after removing the homologous proteins or similar drugs in training data. Nevertheless, HampDTI still outperforms other state-of-the-art methods, achieving ROC-AUC scores greater than 0.85.

Due to the sparsity of the data, DTINet dataset contains some drugs and proteins with only one interaction partner\citep{NeoDTI}, and we name these drugs, proteins and their interactions ‘unique’. Predicting interactions between ‘unique’ drugs and proteins provides a more challenging\citep{van2014biases} way of evaluating the performances of prediction methods. We divide all DTIs(drugs,targets) into unique DTIs(drugs,targets) and non-unique DTIs(drugs,targets), then construct the HampDTI model based on non-unique DTIs and employ it to predict unique DTIs. It worths mentioning that unique DTIs(drugs,targets) are unseen in the training set. As shown in Figure \ref{figcompare} (f), HampDTI outperforms all baseline methods significantly, at least 3\% in PR-AUC and 4.4\% in ROC-AUC, showing good inductive capability. Therefore, HampDTI has the great potential in predicting new DTIs for drugs or targets that do not have much a priori DTI knowledge with better generalization ability. 

\subsubsection{Performances evaluated by independent test}
To evaluate the generalization performance of the models more fairly, we construct the prediction models on the cross-validation set, and make prediction for the independent test set. Table \ref{tabindepent} shows the results on the independent test set. HampDTI produces slightly lower ROC-AUC score and PR-AUC score on the independent test set, but still outperforms other state-of-the-art methods. Although independent test set is not used for the model training and parameter setting, HampDTI performs similarly on the the independent test and the cross-validation set, and the results demonstrate that HampDTI has good generalization capability for unseen data. 

\subsection{Ablation Experiments}

To further investigate the contribution of different components of the HampDTI, we conducted an ablation study by considering several variants:

\begin{figure}[!htbp]
\centering
      \includegraphics[width=0.9\columnwidth]{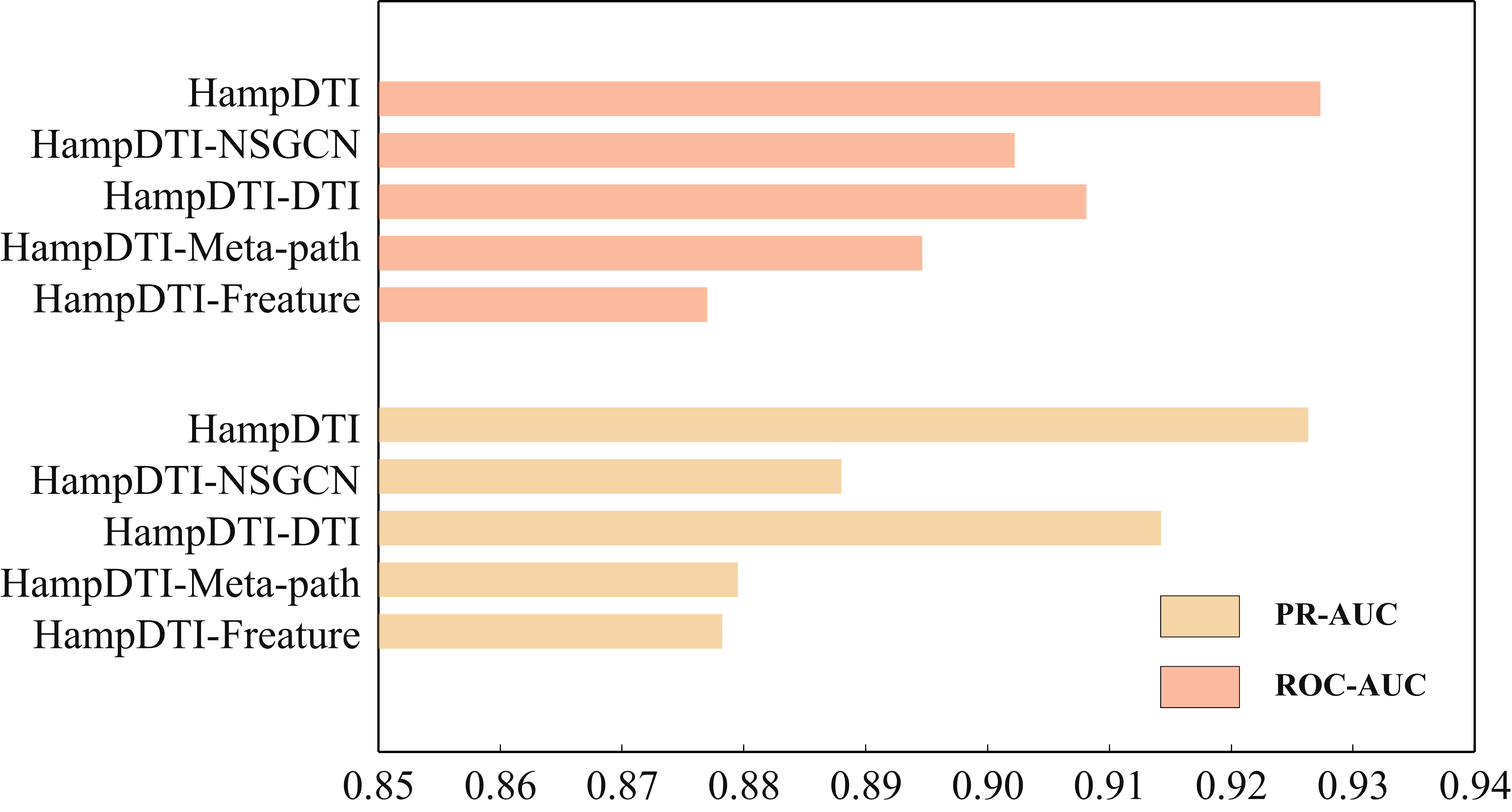}
\caption{The result of the ablation study.}
\label{figablation}
\end{figure}

\begin{itemize}
\item \textbf{HampDTI-Feature}: we remove the feature learning module, and use randomly initialized vectors of the same dimension as node attributes of meta-path graphs.
\item \textbf{HampDTI-Meta-path}: we remove the automatic meta-path learning module and only feed the drug-target interaction graph to the NSGCN module.
\item \textbf{HampDTI-DTI}: we remove the drug-target interaction graph in the NSGCN module and only leverage meta-path graphs.
\item \textbf{HampDTI-NSGCN}: we remove the NSGCN module and use GCN to replace it.
\end{itemize}

All prediction models are evaluated by 10-fold cross-validation on the cross validation set, and results are shown in Figure \ref{figablation}. In general, all variants produce the decreased performances, indicating the critial components are important for ensuring the high-accuracy performances of HampDTI. The comparison between HampDTI and HampDTI-Feature show that relevant node attributes can effectively enhance the performance of graph neural networks compared to random initialization of node features. The results of HampDTI and HampDTI-Meta-path demonstrate that automatic meta-path module can well leverage information from the heterogeneous network and contribut significantly to the DTI prediction. Meta-paths include direct drug-target interactions and the indirect associations between drugs and targets. The results of HampDTI, HampDTI-Meta-path and HampDTI-DTI indicate that the direct and indirect associations in the heterogeneous network are useful. Compared to the GCNs employed by HampDTI-NSGCN on the meta-path graphs, the NSGCN adopted by HampDTI produces much better results.

\section{Discussion}

\begin{figure*}[!htbp]
\centering
\includegraphics[width=1.8\columnwidth]{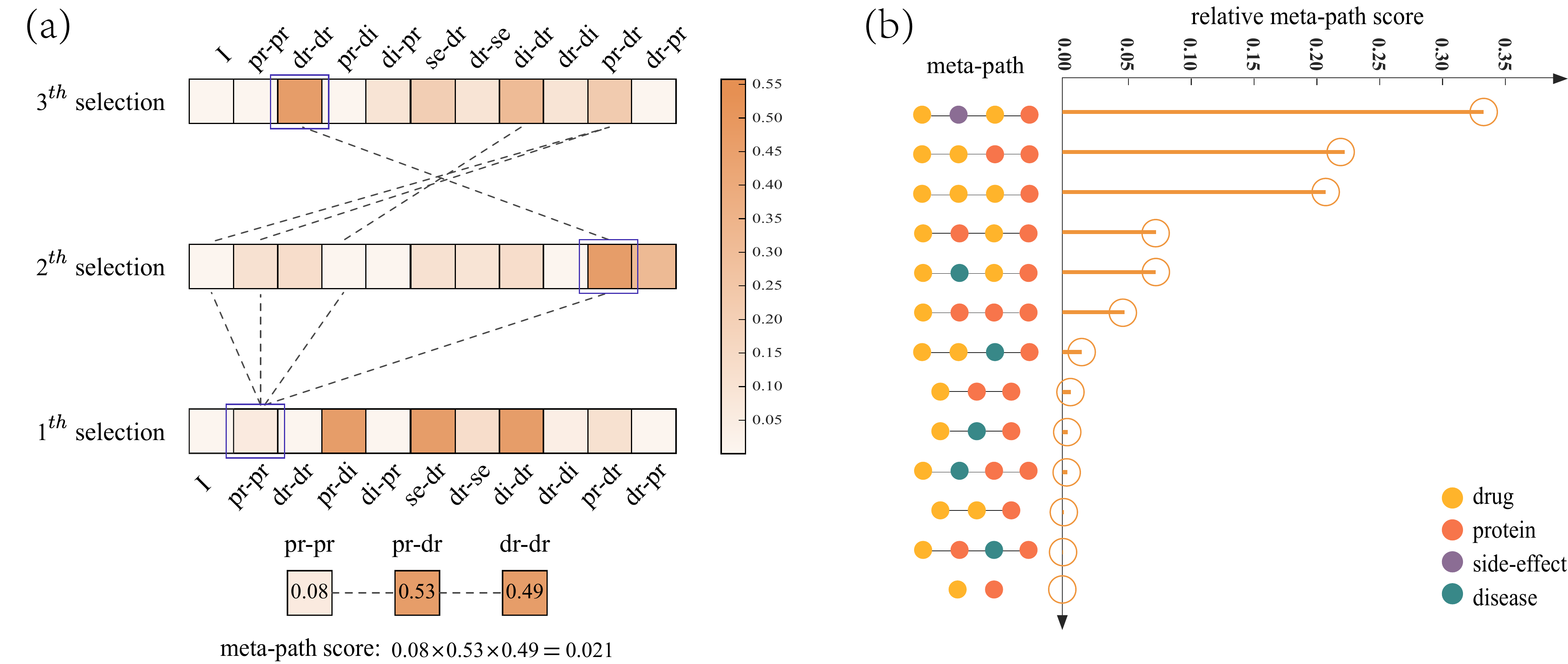}
\caption{Visual presentation of the meta-paths learned by HampDTI. (a)Visualization of the weight $\alpha_k^{(i)}$ in three soft selection. (b)The analysis of relative meta-path scores among all 13 meta-paths in DTINet heterogeneous network.}
\label{figpath}
\end{figure*}

The main contribution of this paper is the design of automatic meta-path learning for the drug and target related heterogeneous networks. Automatic meta-path learning has two advantages. First, it avoids dependency on domain knowledge and can adaptively learn meta-paths between drugs and targets from heterogeneous networks. Second, it provides better interpretability compared to the previous black-box deep learning models, which can reveal some of the most important meta-paths for DTI prediction and deliver some valuable insights to DTI-related studies. 

The interpretability comes from the model parameters $\alpha_k^{(i)}$ in Eq. (\ref{eq04}). As illustrated in Section 2.2.2, the parameters $\alpha_k^{(i)}$ is a weight of relation $R_k$ in $i^{th}$ soft selection. Figure \ref{figpath} (a) visualizes the weight $\alpha_k^{(i)}$ of the trained model in the first fold  of ten-fold cross-validation experiment, where the dashed line shows all four possible meta-paths if the first selection relation is $R_2$(protein-protein interaction). For each possible meta-path $P$ with length $p$, we can see that HampDTI can learn a corresponding sequence of weight $\{ \alpha^{(i)} \}_{i=1}^p$. We define the product of these weights $\prod_{i=1}^{p} \alpha^{(i)}$ as meta-path score, which can measure the importance of the meta-path $P$. Then, we calculate the ratio of the meta-path scores to the sum of all meta-path scores, namely relative meta-path scores, which reflect the contribution of a meta-path in all meta-paths. For example, the meta-path score of the meta-path $P_1: protein \stackrel{e_{1}}{\rightarrow} protein \stackrel{e_{9}}{\rightarrow} drug \stackrel{e_{2}}{\rightarrow} drug$ is equal to $\alpha_1^{(1)} \times \alpha_9^{(2)} \times \alpha_2^{(3)} = 0.021$. For the DTINet heterogeneous network, there are 13 possible meta-paths between drugs and targets of length less than or equal to 3, and the meta-path $P_1$ make the $22.5\%$ contributions. The contributions of all 13 meta-paths are shown in Figure \ref{figpath} (b), and We found that the top 6 most important meta-paths reaches 97\% contribution, indicating that these meta-paths make the most contributions to the DIT prediction. To further demonstrate the effectiveness of learned important meta-paths, we only consider the meta-paths with great relative meta-path scores and remove others to build the HampDTI model. Specifically, we sort the weights $\alpha^{(i)}$ in descending order at each soft selection, keeping the top $N$ weights until their sum reaches 99\% of the sum of the weights, and fix remaining weights to zero. Then, we generate new meta-path graphs with the modified parameter $\alpha^{(i)}$, and build prediction models. HampDTI only using the most important meta-paths produces very close performances compared to HampDTI that use all meta-paths, having the decline less than 0.01 in terms of ROC-AUC and PR-AUC. The results demonstrate that HampDTI can automatically learn the importance of the meta-paths and build the high-accuracy prediction models.

For DTI prediction, some meta-paths proposed in previous studies\citep{fu2016predicting}: the meta-path $drug \rightarrow drug\rightarrow protein$ indicates that drugs with similar chemical structures may interact with the same target protein. The meta-path $drug \rightarrow protein \rightarrow drug \rightarrow protein$ show that if two drugs share the same protein target, they may also share other protein targets. We can see that HampDTI learned some meta-paths (the top 6 meta-paths in Figure \ref{figpath} (b)) consistent with previous studies. In addition, HampDTI also discovered some new meta-paths, i.e., $drug \rightarrow drug \rightarrow protein \rightarrow protein$. These meta-paths contain field-related knowledge, therefore automatically learned meta-paths have the potential to discover novel knowledge from complex heterogeneous graphs or knowledge graphs.

\section{Conclusion}
The heterogeneous network contains multiple bio-entities and relations with enrich information, and benefits drug-target interaction prediction. Exploring important meta-paths between drugs and targets in heterogeneous networks is important for building high-accuracy prediction models. However, meta-paths are usually set by domain experts empirically in previous studies, and cannot be well transferred to different heterogeneous networks and tasks. In this work, we propose an automatic meta-path learning based DTI prediction method HampDTI, which avoids the dependence on domain knowledge and automatically learns meta-paths. The experimental results show that HampDTI outperforms other state-of-the-art DTI prediction methods, and obtains the important meta-paths that provide new perspectives for drug discovery. In the future, we will apply HampDTI to other heterogeneous network-based link prediction tasks, such as drug-drug interactions and drug-disease associations.

\section*{Acknowledgements}

\section*{Funding}
This work was supported by the National Natural Science Foundation of China (62072206, 61772381); Huazhong Agricultural University Scientific \& Technological Self-innovation Foundation; Fundamental Research Funds for the Central Universities (2662021JC008). The funders have no role in study design, data collection, data analysis, data interpretation or writing of the manuscript.\vspace*{-12pt}

\bibliographystyle{natbib}

\bibliography{document}
\end{document}